\def\year{2021}\relax
\documentclass[letterpaper]{article}   
\usepackage{aaai21}    
\usepackage{times}    
\usepackage{helvet}   
\usepackage{courier}    
\usepackage[hyphens]{url}    
\usepackage{graphicx}   
\usepackage{natbib}     
\usepackage{caption}
\usepackage{graphicx} 
\usepackage{multirow}
\usepackage{adjustbox}
\usepackage{amssymb,amsmath}
\usepackage{algorithm}      
\usepackage{algpseudocode}
\usepackage{bm}            
\usepackage{subfigure}
\usepackage{amsfonts}
\usepackage{makecell}

\title{Faster Depth-Adaptive Transformers}

\author{
     Yijin Liu, \textsuperscript{1}\thanks{\ \ This work was done when Yijin Liu was interning at Pattern Recognition Center, WeChat AI, Tencent Inc, China} 
     Fandong Meng,\textsuperscript{2} 
     Jie Zhou,\textsuperscript{2} 
     Yufeng Chen\textsuperscript{1} 
     and Jinan Xu\textsuperscript{1}\thanks{ \ \ Jinan Xu is the corresponding author of the paper.} \\
}
\affiliations{
    \textsuperscript{\rm 1}Beijing Jiaotong University, China  \\
    \textsuperscript{\rm 2}Pattern Recognition Center, WeChat AI, Tencent Inc, China \\
    \{yijinliu, fandongmeng, withtomzhou\}@tencent.com \\
    \{chenyf, jaxu\}@bjtu.edu.cn
}

\begin{document}
\maketitle

\begin{abstract}
Depth-adaptive neural networks can dynamically adjust depths according to the hardness of input words, and thus improve efficiency. 
The main challenge is how to measure such hardness and decide the required depths ({\em i.e.,} layers) to conduct. 
Previous works generally build a halting unit to decide whether the computation should continue or stop at each layer. As there is no specific supervision of depth selection, the halting unit may be under-optimized and inaccurate, which results in suboptimal and unstable performance when modeling sentences.
In this paper, we get rid of the halting unit and estimate the required depths in advance, which yields a faster depth-adaptive model. Specifically, two approaches are proposed to explicitly measure the hardness of input words and estimate corresponding adaptive depth, namely 1) mutual information (MI) based estimation and 2) reconstruction loss based estimation.
We conduct experiments on the text classification task with 24 datasets in various sizes and domains. 
Results confirm that our approaches can speed up the vanilla Transformer (up to 7x) while preserving high accuracy.
Moreover, efficiency and robustness are significantly improved when compared with other depth-adaptive approaches.
\end{abstract}

\section{Introduction}
In the NLP literature, neural networks generally conduct a fixed number of computations over all words in a sentence, regardless of whether they are easy or difficult.
In terms of both efficiency and ease of learning, it is preferable to dynamically vary the numbers of computations according to the hardness of input words \cite{UT_2019}.

\citeauthor{ACT_2016} (\citeyear{ACT_2016}) firstly proposes adaptive computation time (ACT) to improve efficiency of  neural networks.
Specifically, ACT employs a halting unit upon each word when processing a sentence, then this halting unit determines a probability that computation should continue or stop layer-by-layer. 
Its application to sequence processing is attractive and promising.
For instance, ACT has been extended to reduce computations either by exiting early or by skipping layers for the ResNet \cite{ACT_resnet_2017}, the vanilla Transformer \cite{depth_ada_transformer_2020}, and the Universal Transformer \cite{UT_2019}.

However, there is no explicit supervision to directly train the halting unit of ACT, and thus how to measure the hardness of input words and decide required depths is the key point. Given a task, previous works generally treat the loss from different layers as a measure to implicitly estimate the required depths, {\em e.g.,} gradient estimation in ACT, or reinforcement rewards in SkipNet \cite{skipnet_2018}. Unfortunately, these approaches may lead to inaccurate depth selections with high variances, and thus unstable performance.
More recently, the depth-adaptive Transformer \cite{depth_ada_transformer_2020} directly trains the halting unit with the supervision of `pseudo-labels', which is generated by comparing task-specific losses over all layers. Despite its success, the 
depth-adaptive Transformer still relays on a halting unit, which brings additional computing costs for depth predictions, hindering its potential performance.

In this paper, we get rid of a halting unit when building our model, and thus no additional computing costs need to estimate depth. Instead, we propose two approaches to explicitly estimate the required depths in advance, which yield a faster depth-adaptive Transformer.
Specifically, the MI-based approach calculates the mutual dependence between a word and all categorical labels. The larger the MI value of the word is, the more information of labels is obtained through observing this word, thus fewer depths are needed to learn an adequate representation for this word, and vice versa. Due to the MI-based approach is purely conducted in the data preprocessing stage, the computing cost is ignorable when compared with training a neural model in the depth-adaptive Transformer.
The reconstruction loss based approach measures the hardness of learning a word by reconstructing it with its contexts in a sentence. The less reconstruction loss of the word is, the more easily its representation is learned. Therefore the index of the layer with minimum reconstruction loss can be regarded as an approximation for required depths. 
As a by-product, the reconstruction loss based approach is easy to apply to unsupervised scenarios, as it needs no task-specific labels.
Both of the above approaches aim to find a suitable depth estimation. Afterward, the estimated depths are directly used to guide our model to conduct corresponding depth for both training and testing.  

Without loss of generality, we base our model on the Transformer encoder.
Extensive experiments are conducted on the text classification task (24 datasets in various sizes and domains). Results show that our proposed approaches can accelerate the vanilla Transformer up to 7x, while preserving high accuracy. Furthermore, we improve the efficiency and robustness of previous depth-adaptive models.

Our main contributions are as follows\footnote{Codes will appear at https://github.com/Adaxry/Adaptive-Transformer}:

\begin{itemize}
\item We are the first to estimate the adaptive depths in advance and do not rely on a halting unit to predict depths.
\item We propose two effective approaches to explicitly estimate the required computational depths for input words. Specifically, the MI-based approach is computing efficient and the reconstruction loss based one is also applicable in unsupervised scenarios.
\item Both of our approaches can accelerate the vanilla Transformer up to 7x, while preserving high accuracy. Furthermore, we improve previous depth-adaptive models in terms of accuracy, efficiency, and robustness.
\item We provide thorough analyses to offer more insights and elucidate properties of our approaches. 
\end{itemize}

\section{Model}
\subsection{Depth Estimation}
\label{how_to_estimate_sec}
In this section, we introduce how to quantify the hardness of learning representations for input words and obtain corresponding estimated depths.

\paragraph{Mutual Information Based Estimation.}
Mutual Information (MI) is a general concept in information theory. It measures the mutual dependence between two random variables $X$ and $Y$. Formally, the MI value is calculated as:

\begin{equation}
    \mathrm{MI}(X ; Y)= 
    \sum_{y \in Y} \sum_{x \in X} p_{(X, Y)} 
    \cdot \log (\frac{p_{(X, Y)}(x, y)}{p_{X}(x) \cdot 
    p_{Y}(y)} )
\end{equation}    
where $p_{(X,Y)}$ is the joint probability of $X$ and $Y$, and $p_{X}$ and $p_{Y}$ are the probability functions of $X$ and $Y$ respectively.

MI has been widely used for feature selection in the statistic machine learning literature \cite{mi_feature_selection_2005}.
In our case of text classification, $X$ is the set of all words, and $Y$ is the set of predefined labels. Given a word $x \in X$ and a label $y \in Y$, the value of $\mathrm{MI}(x,y)$ measures the degree of dependence between them. The larger $\mathrm{MI}(x,y)$ is, the greater certainty between this word $x$ and label $y$ is, and thus fewer computations are needed to learn an adequate representation for $x$ to predict $y$. For example, the word `terrible' can decide a `negative' label with high confidence in sentiment analysis tasks, and thus it is unnecessary to conduct a very deep transformation when processing words with high MI values, and vice versa. 
Namely, we force our models not to merely focus on a few `important' words and pay more attention to overview contexts when learning the representation of a sentence.
In this way,
our models avoid overfilling limited `important' words, which also takes an effect of regularization, and thus improve generalization and robustness.
Based on the above assumptions, it is intuitive and suitable to choose MI to quantify the difficulty of learning a word.

\begin{figure}[t!]
\begin{center}
     \scalebox{0.46}{
      \includegraphics[width=1\textwidth]{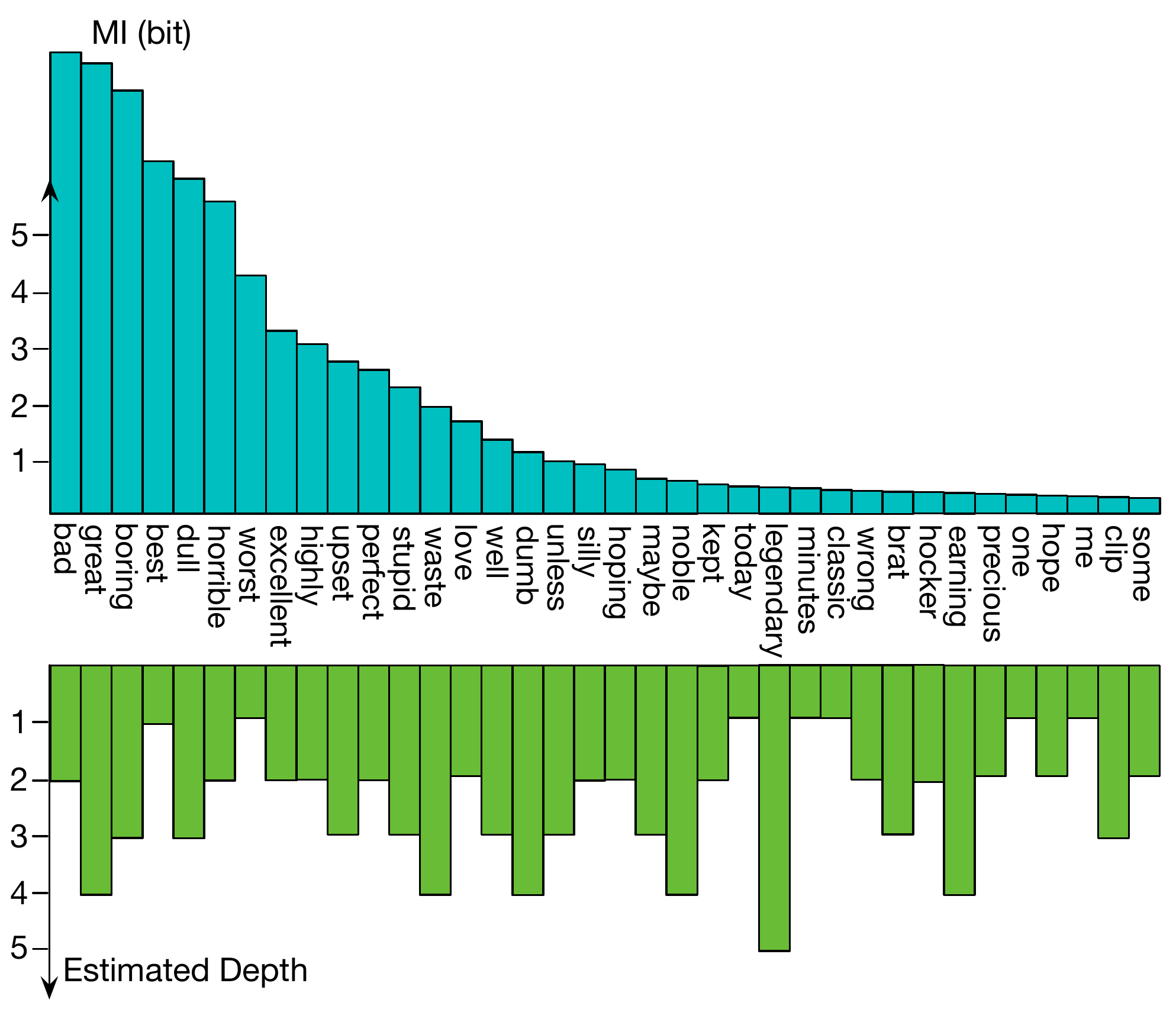}
      }
    
      \caption{
      The histogram of MI values of partial words from IMDB (the upper part), and corresponding depths of these words by using the reconstruction loss based estimation (the bottom part). 
       } \label{long_tail}
       
 \end{center}
\end{figure}

Formally, given a dataset with vocab $X$ and label set $Y$, the MI value $\mathrm{MI}(x)$ for word $x$ is calculated as:

\begin{equation}
\begin{split}
    \mathrm{MI}(x) = & 
    \sum_{y \in\{Y \}}
    \sum_{i_{x} \in\{0,1 \}}
    \sum_{i_{y} \in\{0,1 \}}
    P(i_{x}, i_{y}) \\ & \cdot  \log \left(\frac{
    P\left(i_{x}, i_{y}\right)} {P\left(i_{x}\right) \cdot P\left(i_{y}\right)}
    \right)
\end{split}
\label{eqn_mi_cls}
\end{equation}
where $i_{x}$ is a boolean indicator whether word $x$ exists in a sentence. Similarly, $i_{y}$ indicates the existence of label $y$. 
In practice, the probability formulas $P(\cdot)$ in Equation (\ref{eqn_mi_cls}) are calculated by frequencies of words, labels, or their combinations. A smooth factor (0.1 in our experiments) is introduced to avoid zero division. To avoid injecting information of golden labels when testing, we only use the training set to calculate MI values,

Once the MI value $\mathrm{MI}(x)$ for each word is obtained, we proceed to generate the pesudo-label of depth distribution $d(x)$ accordingly. As the histogram of MI values shown in Figure \ref{long_tail} (the upper part), there is an obvious long tail phenomenon, which manifests that the distribution is extremely imbalanced. To alleviate this issue, we perform a logarithmic scaling for the original $\mathrm{MI}(x)$ as:
\begin{equation}
    \begin{split}
     \operatorname{MI}_{log}(x) = -\log \left(
     \operatorname{MI}(x)
     \right)
    \end{split}
\end{equation}
Next, according to the scaled $\mathrm{MI}_{log}(x)$, we uniformly divide all words into $N$ bins \footnote{We set $N$ to 12 for the compatibility of BERT.}  with fixed-width margin, where $N$ denotes a predefined number of bins ({\em i.e.,} maximum depth). Consequently, the estimated depth value $d(x)$ for word $x$ is the index of corresponding bins.

The MI-based approach is purely calculated at the data preprocessing stage, thus it is highly efficient in computation and does not rely on additional trainable parameters.

\paragraph{Reconstruction Loss Based Estimation.}
Generally in a sentence, several words may bring redundant information that has been included by their contexts. Thus if we mask out these trivial words, it would be easier to reconstruct them than others. Namely, The less reconstruction loss of a word is, the more easily its representation is learned.
Based on the above principle, we utilize this property of reconstruction loss to quantify the hardness of learning the representation for input words and then estimate their required depths.
Firstly, we finetune BERT \cite{bert_2019} with a masked language model task (MLM) on datasets of downstream tasks. Note that we modify BERT to make predictions at any layers with a shared classifier, which is also known as {\em anytime prediction} \cite{huang2017multi,depth_ada_transformer_2020}.
The losses from all layers are summed up \footnote{We experimented with different weights ({\em e.g.,} random sample, or linearly decaying with the number of layers) for different layers, and finally choose the simple equal weights.} to the final loss. 
After finetuning the MLM, given an input sentence $\boldsymbol{x}$ with $|\boldsymbol{x}|$ words, we sequentially replace each word $\boldsymbol{x}_t$ ($t \in [1,|\boldsymbol{x}|]$) with a special symbol {\tt <MASK>}, and then feed the sentence with a {\tt <MASK>} into the MLM. 
Finally, the index of a layer with the minimum loss is selected as the estimated depth value $d(\boldsymbol{x}_t)$: 

\begin{equation}
\label{log_scale_mi}
    \begin{split}
     d(\boldsymbol{x}_t) = \mathop{\arg\min}_{n}(loss_{n} - \lambda n)
    \end{split}
\end{equation}
where $n \in N$ is the index of layer, $loss_{n}$ is the loss of {\tt <MASK>} in the $n$-th layer, and $\lambda n$ is the penalty factor to encourage a lower selection 
\footnote{We elaborate the effect and choice of $\lambda$ in the following analytical Section.}. 
Specifically, we train MLMs following the experimental setup of BERT \cite{bert_2019} with two major differences: 1) We make predictions at every layer with a shared classifier instead of only at the final layer in BERT; 2) We remove the next sentence prediction task following RoBERTa \cite{liu2019roberta}.

\paragraph{Comparisons Between the Two Approaches.}
Although the above approaches perform differently, they both serve as a measure to estimate required depths for input words from the perspective of learning their representations. We proceed to make a detailed comparison between the two approaches. 

In the term of computational cost, the MI-based approach calculates MI values, and then stores the word-depth pairs that resemble word embeddings. The above procedures merely happen at the stage of data preprocessing, which requires trivial computational cost and does not rely on additional trainable parameters, and thus the MI-based approach is highly efficient in computation. In contrast, the reconstruction loss based approach needs to train several MLMs with \textit{anytime prediction}, which yields extra computational costs. Considering the MLMs are dependent on the main model, the calculation of depths can be conducted in advance in a piplined manner.

As the histogram shown in Figure \ref{long_tail}, we observe different preferences between the two estimations. Firstly, the MI-based approach (upper part) tends to assign higher MI values to label-relevant words ({\em e.g.,} opinion words `perfect' and `horrible' in IMDB). After the scaling function described by Equation (3), these opinion words are assigned a lower number of depths, namely fewer computational steps.
Such operations make our models not only focus on a few `important' words, but also pay more attention to the overview contexts, which takes an effect of regularization, and thus improve generalization and robustness.

Unlike the bias for label-related words in the MI-based approach, the reconstruction based approach (bottom part in Figure \ref{long_tail})  purely relies on the unsupervised context to measures the hardness of learning, which is good at recognizing common words ({\em e.g.,} `today', `one' and `me'), and assigns a smaller number of computations, and vice versa. 
As a by-product, the reconstruction loss based approach is applicable to unsupervised scenarios, as it needs no task-specific labels.

\begin{figure}[t!]
\begin{center}
     \scalebox{0.4}{
      \includegraphics[width=1\textwidth]{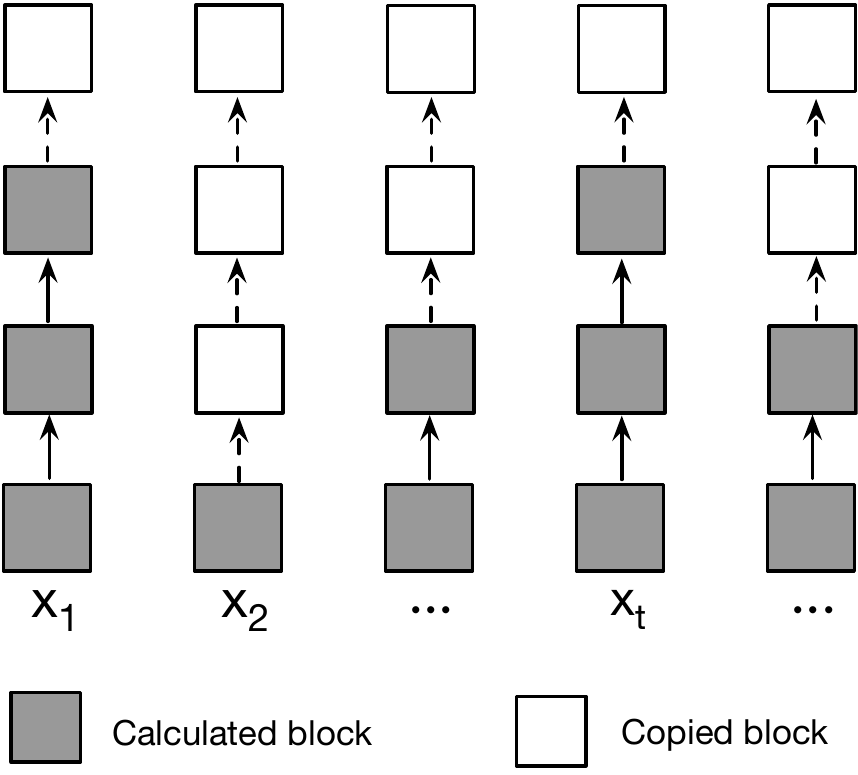}
      }
      
      \caption{The overview of our depth-adaptive Transformer.
      Once a word $\boldsymbol{x}_t$ achieve its own depth $d(\boldsymbol{x}_t)$, it will simply copy states to upper layers. }
      \label{overview}
      
 \end{center}
\end{figure}

\begin{table*}[t!]
\begin{center}
\scalebox{1}{
\begin{tabular}{l|c c c c c c c} 
\hline
\textbf{Dataset} & Classes & Type & \makecell{Average \ \ \\ \ \ Lenghts \ \ \  } & \makecell{\ \ Max \ \ \\ \ \ Lengths} &
\makecell{Train \ \ \\ \ \  Sample \ \ } & \makecell{Test \ \ \\ \ \ Sample \ \ } \\
\hline
TREC  \cite{trec_2002}      & 6        & Question   & 12   & 39       & 5,952   &  500    \\
AG’s News \cite{char_cnn_2015}  & 4        & Topic      & 44   & 221     & 120,000 &  7,600  \\
DBPedia  \cite{char_cnn_2015}     & 14       & Topic      & 67   & 3,841   & 560,000 &  70,000 \\
Subj \cite{subj_2004}        & 2        & Sentiment  & 26   & 122      & 10,000  &  CV   \\
MR \cite{MR_2005}         & 2        & Sentiment  & 23   & 61       & 10,622  &  CV   \\
Amazon-16 \cite{16_cls_data_17}   & 2        & Sentiment  & 133  & 5,942   & 31,880  &  6,400 \\
IMDB \cite{IMDB_2011}        & 2        & Sentiment  & 230  & 2,472   & 25,000  &  25,000 \\  
Yelp Polarity \cite{char_cnn_2015} & 2        & Sentiment  & 177  & 2,066    & 560,000 &  38,000 \\ 
Yelp Full \cite{char_cnn_2015} & 5  & Sentiment  & 179  & 2,342    & 650,000 &  50,000 \\
\hline
\end{tabular}}
\end{center}

\caption{Dataset statistics. `CV' refers to 5-fold cross-validation. There are 16 subsets in Amazon-16.
}

\label{data_statistics}
\end{table*}

\subsection{Depth-Adaptive Mechanism}
As the overview shown in Figure \ref{overview}, we stack $N$ layers of the Transformer encoder to model a sentence. The Transformer encoder consists of two sub-layers in each layer. The first sub-layer is a multi-head dot-product self-attention and the second one is a position-wise fully connected feed-forward network. We refer readers to the original paper \cite{Transformer_2017} for more details.

To make sure all hidden states of the same layer are available to compute self-attention, once a word $\boldsymbol{x}_t$ reaches its own maximal layer $d(\boldsymbol{x}_t)$, it will stop computation, and simply copy its states to the next layer until all words stop or the 
maximal layer $N$ is reached. Formally, at the $n$-th layer, for the word $x_t$, its hidden state $\boldsymbol{h}_i^{n}$ are updated as follows:

\begin{equation}
\boldsymbol{h}_t^{n} =
\begin{cases}
\boldsymbol{h}_{t}^{n-1} \text{ \ \ \ \ \ \ \ \ \ \ \ \ \ \ \ \ \ \ \ \ \ \ \ \ \ \ \ \ \ \ \ \ \ if $ n > d(x_t) $}  \\
\operatorname{Transformer}( \boldsymbol{h}_{t}^{n-1})  \text{\ \ \ \ \ \ \ \ \ \ else} 
\end{cases}
\end{equation}
where $n \in [1, N]$ refers to the index of the layer.
Especially, $\boldsymbol{h}_{t}^{0}$ is initialized by the BERT embedding.

\subsection{Task-specific Settings}
After dynamic steps of computation for each word position, we make task-specific predictions upon the maximal stop layer $n_{max} \in [1, N]$ among all word positions. The feature vector $\boldsymbol{v}$ consists of mean and max pooling of output hidden states $\boldsymbol{h}^{n_{max}}$, and is activated by ReLU. Finally, a softmax classifier are built on $\boldsymbol{v}$. Formally, the above-mentioned procedures are computed as follows:

\begin{equation}
    \begin{split}
        & \boldsymbol{v} = \operatorname{ReLU}([\mathop{\max}(\boldsymbol{h}^{n_{max}}); \operatorname{mean} ({\boldsymbol{h}}^{n_{max}})]) \\
        & P(\widetilde{y} | \boldsymbol{v}) = \operatorname{softmax}(\boldsymbol{W} \boldsymbol{v} + \boldsymbol{b})
        \label{equ_cls_pred}
    \end{split}
\end{equation}
where $\boldsymbol{W} \in \mathbb{R}^{d_{model} \times |S|}$ and $\boldsymbol{b} \in \mathbb{R}^{|S|}$ are parameters of the classifier, $|S|$ is the size of the label set, and $P(\widetilde{y} | \boldsymbol{v})$ is the probability distribution. 
At the training stage, we use the cross-entropy loss computed as:

\begin{equation}
    \begin{split}
    & Loss = - \sum\limits_{i=1}^{|S|}{y_i} log(P_i(\widetilde{y} | \boldsymbol{v})) 
    \end{split}
    \label{equ_task_loss}
\end{equation}
where $y_i$  is the golden label.
For testing, the most probable label $\hat{y}$ is chosen from above probability distribution described by Equation (\ref{equ_cls_pred}):
\begin{equation}
    \begin{split}
        & \hat{y} = \mathop{\arg\max} P (\widetilde{y} | \boldsymbol{v} ) \\
    \end{split}
    \label{background_intent_pred}
\end{equation}

\section{Experiments}

\subsection{Task and Datasets}
Text classification aims to assign a predefined label to text \cite{char_cnn_2015}, which is a classic task for natural language processing and is generally evaluated by accuracy score. Generally, The number of labels may range from two to more, which corresponds to binary and fine-grained classification. We conduct extensive experiments on the 24 popular benchmarks collected from diverse domains ({\em e.g.,} \textit{topic}, \textit{sentiment}) ranging from modestly sized to large-scaled. The statistics of these datasets are listed in Table \ref{data_statistics}.

\begin{table*}[t!]
\begin{center}
\scalebox{0.85}{
\begin{tabular}{l|c c c c c c c}
\hline
Data / Model & \makecell{Multi-Scale \ \ \\ \ \ Transformer \ \ \  } & \makecell{Star- \ \ \\ \ \ Transformer \ \ \  } & Transformer & \makecell{Transformer \ \ \\ \ \ w/ halting unit \ \ \  }   & \makecell{Transformer \ \ \\ \ \ w/ MI estimation \ \ \  } & \makecell{Transformer \ \ \\ \ \ w/ reconstruction  \ \ \  } \\
\hline
Apparel      & 86.5   & 88.7  & \textbf{91.9}  & 91.6  & 91.4  & 91.8 \\ 
Baby         & 86.3   & 88.0  & 88.8  & 88.1  & \textbf{90.6}  & 88.4 \\
Books        & 87.8   & 86.9  & 89.5  & 88.3  & \textbf{89.6}  & 89.5 \\
Camera       & 89.5   & 91.8  & 91.8  & 92.7  & \textbf{93.8}  & 92.9 \\
Dvd          & 86.5   & 87.4  & 88.3  & \textbf{91.7}  & 91.4  & 91.5 \\
Electronics  & 84.3   & 87.2  & \textbf{90.8}  & 89.3  & 90.6  & 90.2 \\
Health       & 86.8   & 89.1  & \textbf{91.7}  & 91.3  & 91.6  & 88.4 \\
Imdb         & 85.0   & 85.0  & 88.3  & 89.8  & 89.5   & \textbf{90.6} \\
Kitchen      & 85.8   & 86.0  & 87.6  & 88.1  & 89.2  & 86.8 \\
Magazines    & 91.8   & 91.8  & 94.2  & \textbf{94.8}  & 94.6  & 94.7 \\
Mr           & 78.3   & 79.0  & \textbf{83.7}  & 81.6  & 82.3  & 82.5 \\
Music        & 81.5   & 84.7  & \textbf{89.9}  & 89.3  & 89.5  & 87.2 \\
Software     & 87.3   & 90.9  & 91.2  & 92.9  & 92.3  & \textbf{93.8} \\
Sports       & 85.5   & 86.8  & 87.1  & 89.2  & 88.4  & \textbf{89.8} \\
Toys         & 87.8   & 85.5  & 89.7  & 90.3  & \textbf{90.9}  & 89.7 \\
Video        & 88.4   & 89.3  & 93.4  & \textbf{94.3}  & 93.1  & 93.5 \\
\hline
Avg          & 86.2   & 87.4  & 89.9  & 90.2  & \textbf{90.5}  & 90.1  \\
\hline
 \end{tabular}}
\end{center} 
\caption{Accuracy scores (\%) on the Amazon-16 datasets. Best results on each dataset are bold. The results of Multi-Scale Transformer \cite{ms_transformer} is cited from the original paper, and other results are our implementations with several recent advanced techniques ({\em e.g.,} BERT initialization) under the unified setting.
}  
\label{amazon16_result}
\end{table*}

\subsection{Implementation Details}
For the MI-based estimation approach, we calculate word-depth pairs on the training set in advance and then calculate depths for words in the test set. For the reconstruction based approach, we calculate word-depth pairs for both train and test set without using label information.
The penalty factor $\lambda $ in the reconstruction loss based approach is set to 0.1.
Dropout \cite{dropout_2014} is applied to word embeddings, residual connection , and attention scores with a rate of 0.1. Models are optimized by the Adam optimizer \cite{Adam_2014} with gradient clipping of 5 \cite{gradient_clip_2013}. 
 $BERT_{base}$ is used to initialize the Transformer encoder. Long sentences exceed 512 words are clipped. 

\subsection{Main Results} 

\begin{table*}[t!]
\begin{center}
\scalebox{0.9}{
\begin{tabular}{l|c c c c c c c c |c}
\hline
{Models / Dataset}  & TREC  & MR & Subj   &  IMDB & AG. &  DBP.  & Yelp P. & Yelp F. & Avg.\\
\hline
RCRN \cite{RCRN_2018}                 & 96.20  & --     & --    & 92.80  & --    & --     & --      & -- & --     \\
Cove  \cite{Cove_2017}                & 95.80  & --     & --    & 91.80  & --    & --     & --      & -- & --     \\
Text-CNN \cite{textcnn_2014}          & 93.60  & 81.50  & 93.40 & --     & --    & --     & --      & -- & --     \\
Multi-QT \cite{MR_efficient_2018}     & 92.80  & 82.40  & 94.80 & --     & --    & --     & --      & -- & --     \\
AdaSent \cite{subj_self_2015}         & 92.40  & 83.10  & 95.50 & --     & --    & --     & --  & --    & --      \\
CNN-MCFA \cite{subj_trans_2018}       & 94.20  & 81.80  & 94.40 & --     & --    & --     & --      & --      & -- \\
Capsule-B \cite{capsule_2018}         & 92.80     & 82.30     & 93.80    & --     & 92.60       & -- & -- & --  & -- \\ 
DNC+CUW \cite{less_memory_2019}       & --     & --     & --    & --   &  93.90  & -- & 96.40 & 65.60 & -- \\
Region-Emb \cite{region_emb_2018}   & --     & --     & --    & --  & 92.80 & 98.90 & 96.40 & 64.90 & -- \\
Char-CNN \cite{char_cnn_2015}         & --     & --     & --    & --     & 90.49 & 98.45  & 95.12   & 62.05   & -- \\
DPCNN \cite{DPCNN_2017} & --     & --     & --    & --  & 93.13  & 99.12  &  97.36 & 69.42 & -- \\
DRNN \cite{DRNN_2018}                 & --     & --     & --    & --     & 94.47 & 99.19  & 97.27   & 69.15   & -- \\
SWEM-concat \cite{swem_2018} & 92.20 & 78.20 & 93.00 & -- & 92.66 & 98.57 & 95.81 & 63.79 & -- \\
Star-Transformer \cite{star_transformer_2019} $\dagger$         & 93.00  & 79.76  & 93.40 & 94.52  & 92.50 & 98.62  & 94.20   & 63.21 & 88.65  \\
BERT \cite{bert_2019} & -- & -- & -- & 95.49 & -- & 99.36 & 98.11 & 70.68 & -- \\
XLNet \cite{XLNet_2019} & -- & -- & -- & \textbf{96.80} & 95.55 & \textbf{99.40} & \textbf{98.63} & 72.95 & -- \\
\hline
Transformer  \cite{Transformer_2017}  $\dagger$  & 96.00  & 83.75  & 96.00 & 95.58  & 95.13 & 99.22  & 98.09  & 69.80 & 91.69 \\
\ \ \ \ w/ Halting unit \cite{depth_ada_transformer_2020} $\dagger$  & 95.80  & 83.23  & 96.00 & 95.80  & 95.50 & 99.30  & 98.25   & 69.75 & 91.70 \\
\ \ \ \ w/ MI estimation (ours) $\dagger$    & \textbf{96.50}  & \textbf{84.20}  & 96.00 & 96.72  & \textbf{95.90} & 99.32  & 98.10  & \textbf{72.98} & \textbf{92.46} \\  
\ \ \ \ w/ Reconstruction estimation (ours)  $\dagger$    & 96.20  & 83.90 & \textbf{96.30} & 96.60  & 95.65 & 99.25  & 98.00  & 69.58 & 91.93 \\                            
\hline
\end{tabular}}
\end{center}
\caption{Accuracy scores (\%) on modestly sized and large-scaled datasets. `AG.',  `DBP.', `Yelp P.' and `Yelp F.' are the abbreviations of `AG's News`, `DBPedia', `Yelp Polarity' and `Yelp Full', respectively. $\dagger$ is our implementations with several recent advanced techniques and \textit{analogous} parameter sizes. `Transformer' is the Transformer encoder initialized by $BERT_{base}$ with 12 fixed layers. 
} 
\label{cls_result}
\end{table*}

\begin{figure*}[t!]
\begin{center}
     \scalebox{1}{
      \includegraphics[width=1\textwidth]{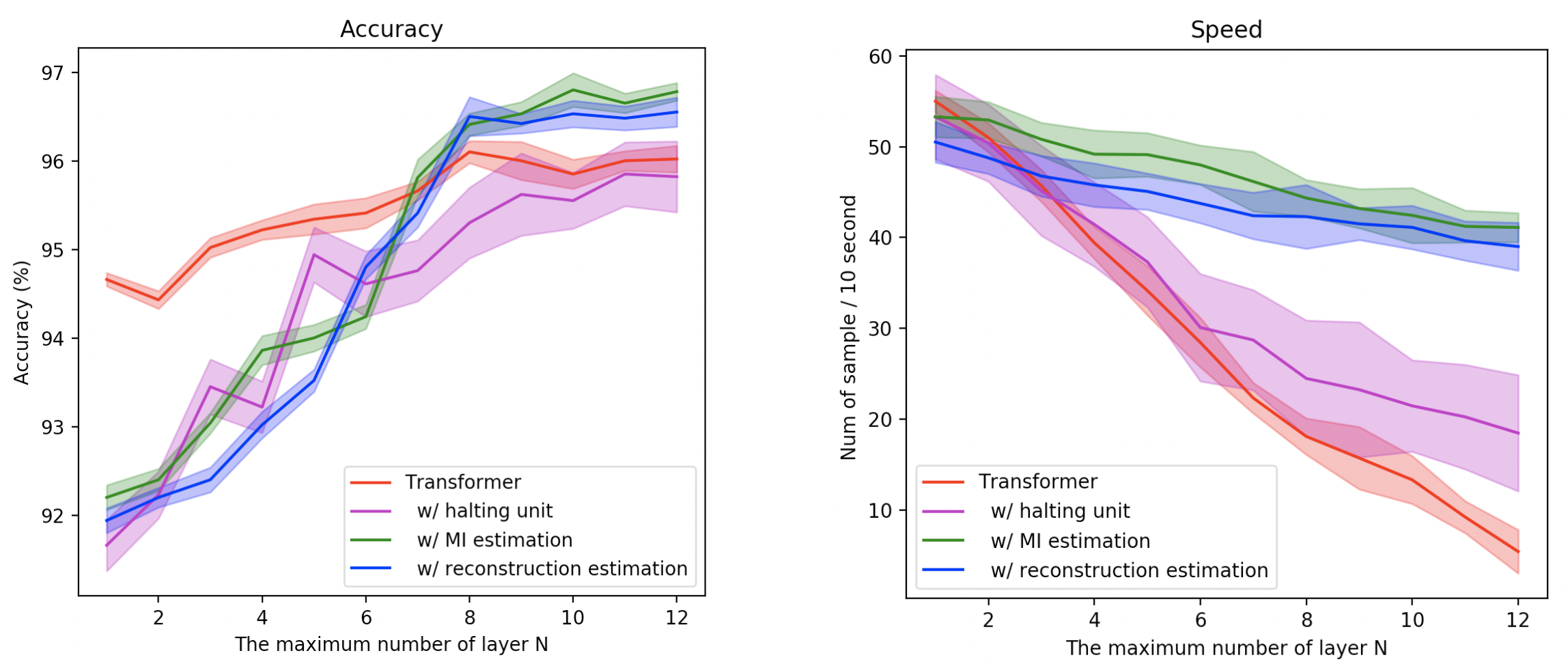}
      }
      \caption{Accuracy scores (a) and speed (b) for each model on IMDB when $N \in [1,12]$. The solid line indicates the mean performance, and the size of the colored area indicates variance (used to measure robustness). `speed' is the number of samples calculated in ten-second on one Tesla P40 GPU with the batch size of 1.  
      } \label{fig_acc_speed} 
\end{center}
\end{figure*}

\paragraph{Results on Amazon-16.}
Amazon-16 consists of consumer comments from 16 different domains ({\em e.g.,} Apparel). 
We compare our approaches with different baseline models in Table \ref{amazon16_result}. The Multi-Scale Transformer \cite{ms_transformer} is designed to capture features from different scales, and the Star-Transformer \cite{star_transformer_2019} is a lightweight Transformer with a star-shaped topology. Due to the absence of a powerful contextual model ({\em e.g.,} BERT), their results underperform others by a margin. The Transformer model is finetuned on BERT and conducts fixed 12 layers for every instance, which yields a strong baseline model. Following the setup of depth-adaptive Transformer, we add a halting unit on the bottom layer of the Transformer encoder, and generate a `pesudo-label' for the halting unit by classification accuracy. Our approaches (the last two columns in Table   \ref{amazon16_result}) achieve better or comparable performance over these strong baseline models. The MI-based approach also takes a regularization effect, and thus it achieves better performance than the reconstruction counterpart.

\paragraph{Results on Larger Benchmarks.}
Although the Amazon-16 benchmark is challenging, its small data size makes the results prone to be unstable, therefore we conduct experiments on larger benchmarks for a more convincing conclusion. In this paragraph, we only focus on the classification accuracy listed in Table \ref{cls_result}, and more detailed results about computing speed and model robustness will be discussed in the next section.

The upper part of Table \ref{cls_result} lists several high-performance baseline models. Their detailed descriptions are omitted here. In terms of accuracy, our approaches achieve comparable performance with these state-of-the-art models. 
At the bottom part of Table \ref{cls_result}, we finetune BERT as our strong baseline model. Results show that this baseline model performs on par with the state-of-the-art XLNet \cite{XLNet_2019}. Then we build a halting unit at the bottom of the baseline model under the same setup with the depth-adaptive Transformer. Results show that applying a halting unit has no obvious impact on accuracy. The last two rows list results of our estimation approaches, where the MI-based approach brings in consistent improvements over the baseline and the Transformer w/ a halting unit, by +0.77\% and +0.76\% on average, respectively. We speculate the improvements mainly come from the additional deep supervision and regularization effect of the MI-based approach. In contrast, the reconstruction based approach only show improvements over the baseline model (+0.24\%) and the Transformer w/ a halting unit (+0.23\%) by a small margin.

\section{Analysis}
We conduct analytical experiments on the modestly sized IMDB to offer more insights and elucidate the properties of our approaches.

\subsection{Effect of the maximum number of layer}
\label{speed_acc_section}
Firstly, we train several fixed-layer Transformers with $N$ ranging from one to twelve, and then build a halting unit on the above Transformers to dynamically adjust the actual number of layers to conduct. Meanwhile, we respectively utilize our two approaches on the fixed-layer Transformer to activate dynamic layers.  Note that each model is trained with different random initialization three times and we report the mean and variance. Here, we take the variance value to measure the robustness against the random initialization and different depth selections.  As drawn in Figure \ref{fig_acc_speed}, solid lines are the mean performance, and the size of the colored areas indicate variances.

\begin{figure}[t!]
\begin{center}
     \scalebox{0.45}{
      \includegraphics[width=1\textwidth]{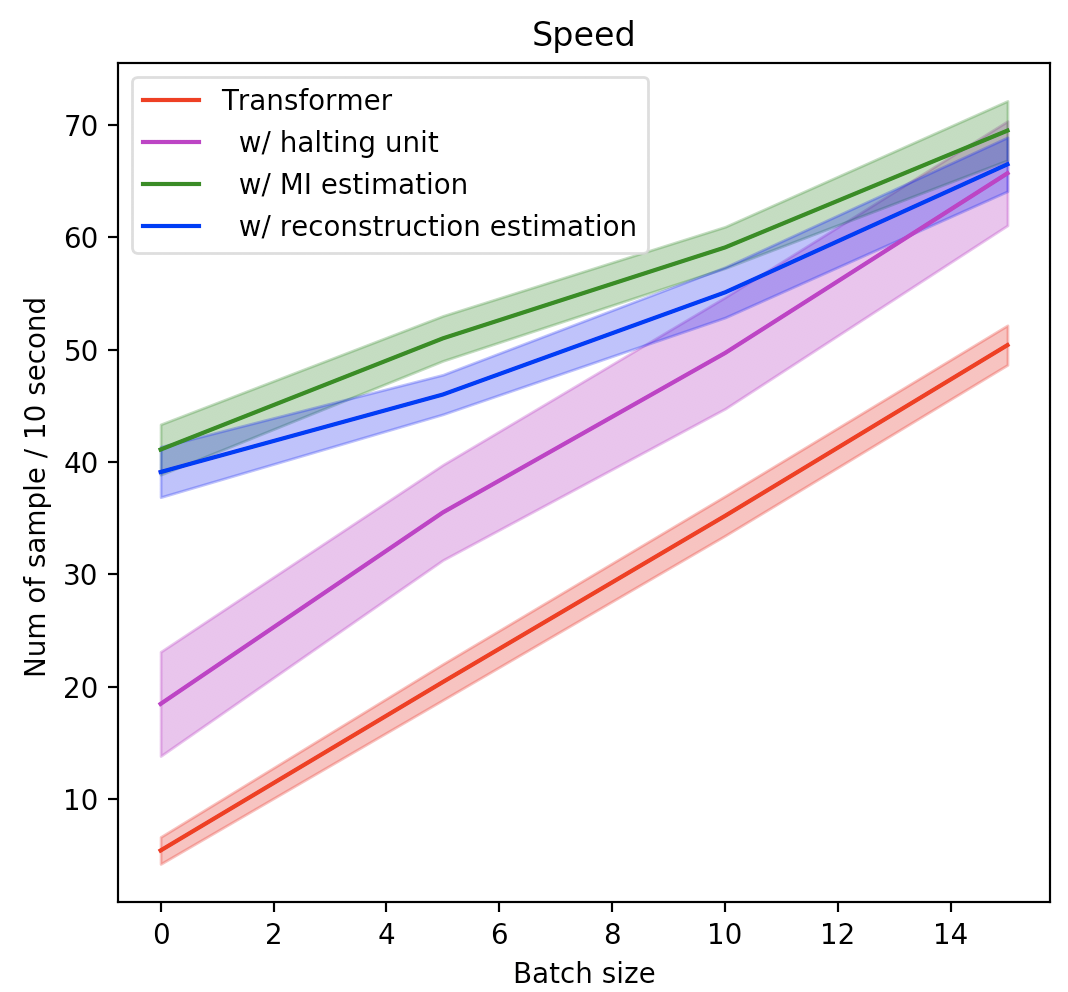}
      }
      \caption{
      Speed for each model on IMDB when $batch size \in [1,15]$. The solid line indicates the mean performance, and the size of the colored area indicates variance (used to measure robustness). `speed' is the number of samples calculated in ten seconds on one Tesla P40 GPU.
       } \label{speed_batch}
       
 \end{center}
\end{figure}

\paragraph{Accuracy and Robustness.}
Results of accuracy and robustness are drawn in Figure \ref{fig_acc_speed} (a). In the lower layers ($N \in [1,6]$), as the searching space for depth selection is small, the depth-adaptive models perform worse than the Transformer baseline. In contrast, when $N \in [6,12]$, the depth-adaptive models come up with the baseline. Due to the additional depth supervision and the regularization effect, the application of our approaches can further significantly improve accuracy and robustness over both the Transformer and w/ a halting unit. (green and blue lines vs. purple line in Figure \ref{fig_acc_speed} (a))

\paragraph{Speed and Robustness.}
Figure \ref{fig_acc_speed} (b) shows the speed and robustness of each model. The speed of vanilla Transformer almost linearly decays with the growth of the number of layers.  As $N \in [1,3]$, due to the additional prediction for depths, models w/ halting unit runs a bit slower than the baseline. However, the superiority of adaptive depths becomes apparent with the growth of the number of layers. In particular, as $N = 12$, the model w/ halting run 3.4x faster than the fixed-layer baseline (pure lines vs. red line in Figure \ref{fig_acc_speed} (b)). As our approaches free the dependency for depth prediction and can further speed up the model, both of our approaches run about 7x faster than the fixed-layer baseline (green and blue lines vs. red line in Figure \ref{fig_acc_speed} (b)).  In addition, our approaches perform more robust in the term of speed gains than the Transformer w/ a halting unit.

\begin{figure}[t!]
\begin{center}
     \scalebox{0.45}{
      \includegraphics[width=1\textwidth]{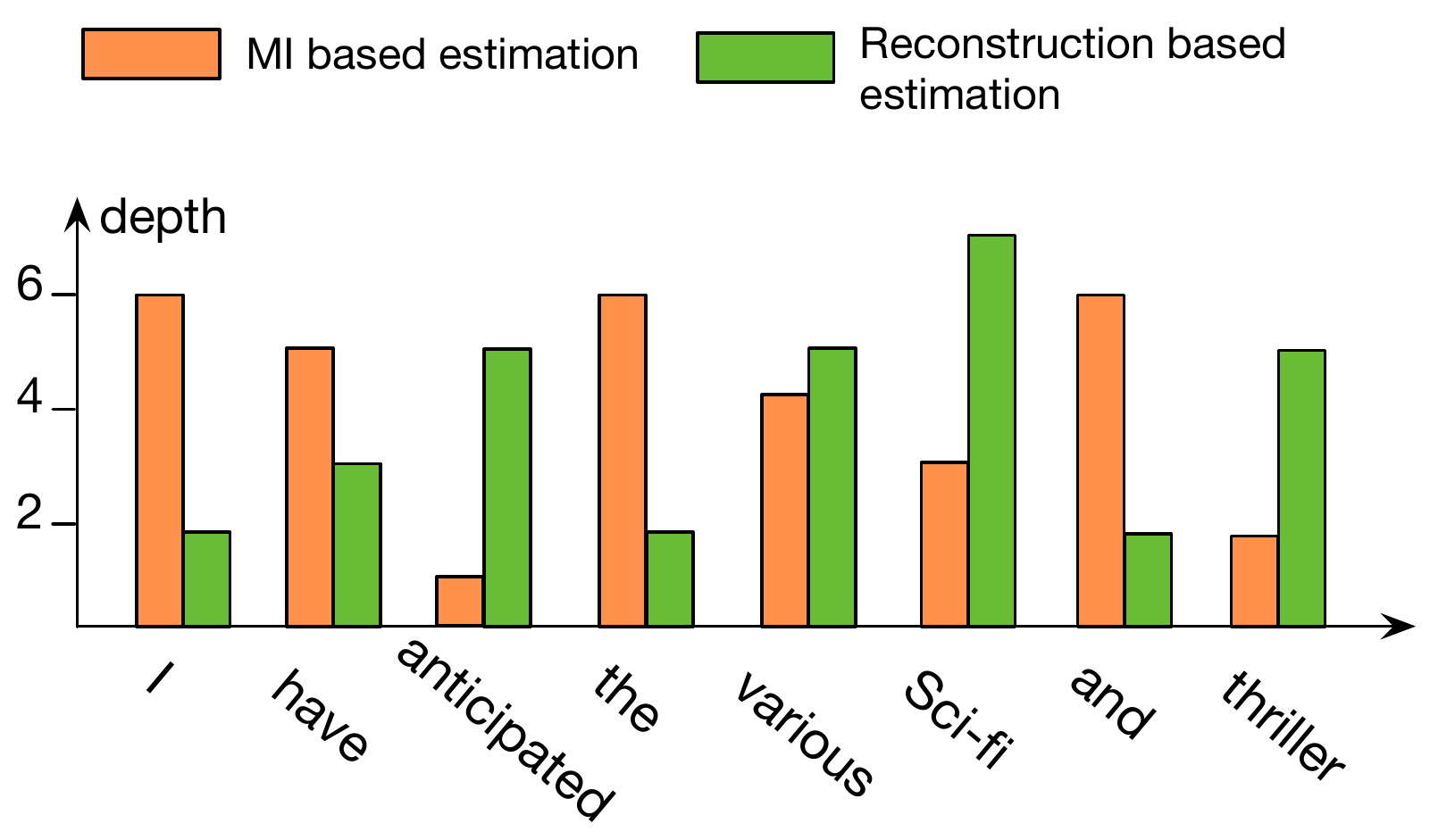}
      } 
      \caption{
       The histogram of the depth distribution of a case from IMDB, which is estimated by our approaches.
       } \label{case}
 \end{center}
\end{figure}

\subsection{Speed on different batch size}
The depth-adaptive models conduct dynamic computations at each word position, and thus the actually activated depths are decided by the maximal depth value. As a result, when the batch size gets larger, the final activated depth may potentially become larger as well, which may hurt the effectiveness of depth-adaptive models. In this section, we fix the maximal number of layer $N$ to 12, and then compare the speed of each model. As shown in Figure \ref{speed_batch}, the speed gain of the depth-adaptive models (green, blue and purple lines in Figure \ref{speed_batch}) grows slower than the vanilla Transformer (red line in Figure \ref{speed_batch}). However, the absolute speed of depth-adaptive models is still much faster than the vanilla Transformer.  We leave the further improvement of depth-adaptive models on larger batch sizes to future works.

\subsection{Effect of penalty factor $\lambda$}
\label{sec_effect_of_lambda}
If no constraints are applied on the depth selection, the reconstruction loss based approach tends to choose a layer as deep as possible, and thus an extra penalty factor $\lambda$ is necessary to encourage a lower choice. We simply search $\lambda \in [0,0.2]$, and finally set it to 0.1 for a good accuracy-speed trade-off. 
The detailed results are list in Table \ref{effect_of_lambda}.

\subsection{Case Study}
We choose a random sentence from the IMDB dataset, and show the estimated depths outputted by both approaches in Figure \ref{case} (upper part). We observe that the MI-based estimation tends to assign a smaller number of depths for opinion words, {\em e.g.,} `anticipated' and `thriller'. While the reconstruction loss based estimation is prone to omit common words, e.g., `and'. 

\begin{table}[t!]
\begin{center}
\scalebox{0.95}{
\begin{tabular}{l|c c c c c c c} 
\hline
$\lambda$     &   0   & 0.05  & 0.10  & 0.15  & 0.20  \\
\hline
accuracy      & 96.54 & 96.31 & 96.55 & 96.27 & 96.29 \\
speed         &  23   & 33    & 48    & 54    & 58    \\
average depth & 9.5   & 6.3   & 4.5   & 3.9   & 3.6   \\
\hline
\end{tabular}}
\end{center}
\caption{Effect of penalty factor $\lambda$. The definition of `speed' is same as that in Figure \ref{fig_acc_speed}. `average depth' is the average predicted depth of words in test set.}
\label{effect_of_lambda}
\end{table}

\section{Related Work}
Our work is mainly inspired by ACT \cite{ACT_2016}, and we further explicitly train the halting union with the supervision of estimated depths. Unlike Universal Transformer \cite{UT_2019} iteratively applies ACT on the same layer, we dynamically adjust the amount of both computation and model capacity. 

A closely related work named `Depth-Adaptive Transformer' \cite{depth_ada_transformer_2020} uses task-specific loss as an estimation of depth selection. 
Our approaches are different from it in three major aspects: 1) We get rid of the halting unit and remove the additional computing cost for depths, thus yield a faster depth-adaptive Transformer; 2) our MI-based estimation does not need to train an extra module, and is highly efficient in computation; 3) our reconstruction loss based estimation is unsupervised, and can be easily applied on general unlabeled texts. Another group of works also aims to improve efficiency of neural network through reducing the entire layers, {\em e.g.,} DynaBERT \cite{dynabert_2020}, LayerDrop \cite{layerdrop_2019} and MobileBERT \cite{mobilebert_2020}. In contrast, our approaches perform adaptive depths in the fine-grained word level.

\section{Conclusion}
We get rid of the halting unit and remove the additional computing cost for depths, thus yield a faster depth-adaptive Transformer.   Specifically, we propose two effective approaches 1) mutual information based estimation and 2) reconstruction loss based estimation. 
Experimental results confirm that our approaches can speed up the vanilla Transformer (up to 7x) while preserving high accuracy.
Moreover, we significantly improve previous depth-adaptive models in terms of accuracy, efficiency, and robustness.
We will further explore the potential improvement of the depth-adaptive Transformer when facing larger batch size in future work.

\section{Acknowledgments.}
The research work described in this paper has been supported by the National Key R\&D Program of China (2020AAA0108001) and the National Nature Science Foundation of China (No. 61976015, 61976016, 61876198 and  61370130). The authors would like to thank the anonymous reviewers for their valuable comments and suggestions to improve this paper.
\bibliography{aaai21}

\end{document}